\documentclass[10pt, a4paper]{article}
\usepackage{lrec}
\usepackage{multibib}
\newcites{languageresource}{Language Resources}
\usepackage{graphicx}
\usepackage{tabularx}
\usepackage{soul}

\usepackage{epstopdf}
\usepackage[latin1]{inputenc}
\usepackage{footnote}
\usepackage{hyperref}
\usepackage{xstring}
\usepackage{tablefootnote}

\title{TAP-DLND 1.0 : A Corpus for Document Level Novelty Detection}

\name{Tirthankar Ghosal, Amitra Salam, Swati Tiwari, Asif Ekbal, Pushpak Bhattacharyya}

\address{Indian Institute of Technology Patna \\
         Bihta, Bihar-801106, India \\
         (tirthankar.pcs16, salam.recs, stlex.re, asif, pb)@iitp.ac.in \\}

\abstract{
Detecting novelty of an entire document is an Artificial Intelligence (AI) frontier problem that has widespread NLP applications, such as extractive document summarization, tracking development of news events, predicting impact of scholarly articles, etc. Important though the problem is, we are unaware of any benchmark document level data that correctly addresses 
the evaluation of automatic novelty detection techniques in a classification framework. To bridge this gap, we present here a resource for benchmarking the techniques for    
\textit{document level novelty detection}. 
We create the resource via \textit{event-specific crawling} of news documents across several domains in a periodic manner. We release the annotated corpus with necessary statistics and show its use with a developed system for the problem in concern.  \\ \newline  }

\begin{document}

\maketitleabstract

\section{Introduction}
Novelty detection implies finding elements that have not appeared before, or new, or original with respect to relevant references. The explosive growth of documents across the web has resulted in the accumulation of redundant ones, thereby consuming space as well as precious time of readers seeking new information. This necessitates finding means for discarding redundant document(s) and retaining ones containing novel information. The level of information duplication is not just limited to the lexical surface form of texts but has encroached the barriers of semantics and pragmatics too. Paraphrasing, semantic level plagiarism etc. are instances of such practices. Intelligent text reuse, synonym replacement and careful alignment may lead to a surface form which is very different from the originating source yet convey the same meaning. Present state-of-the-art text matching techniques are unable to process such redundancy. The quest of new information is an eternal human need and urges attention in this very age of exploratory data redundancy. One major objective of this work is to provide a benchmark setup for experiments to filter out superfluous information across the web. With this work we introduce a simplistic dataset to the research community so as to inculcate efficient methods for detecting \textit{document level novelty} or on the contrary document level redundancy. 
We create the resource by crawling news reporting of events of different categories and coin it as \textit{TAP-DLND 1.0}\footnote{http://www.iitp.ac.in/~ai-nlp-ml/resources.html} (after the initial names of the principal investigators \textit{Tirthankar-Asif-Pushpak}) which also stands for \textit{Explore Document Level Novelty Detection (DLND)}.  In this work we view the problem of novelty detection as a two-class classification problem with the judgment that whether an incoming document bears sufficiently new information to be labeled as novel with respect to a set of source documents. The source document set could be seen as the memory of the reader which stores known information. We extract features from target documents with respect to corresponding source documents and develop a classification system. We report promising results with our features on the developed dataset.
\subsection{Related Works} Although sentence level novelty detection is a well studied problem in information retrieval literature, very little has been done to address the problem at the document level. To begin with
\cite{li2005novelty} rightly pointed out that, research in novelty detection from texts has been carried out at three levels : event level, sentence level and document level.
Research in novelty mining could be traced back to the Topic Detection and Tracking (TDT) ~\cite{allan2002introduction} evaluation campaigns where the concern was to detect new event from online news streams. Although the intention was to detect the \textit{first story} or reporting of a new event from a series of news stories, the notion of \textit{novelty detection} from texts came into light for the research community. 
Some notable approaches for New Event Detection with the TDT corpus are by \cite{allan1998line,yang2002topic,stokes2001first,franz2001first,yang1998study,allan2000first,brants2003system}. However, the Novelty track in TREC \cite{soboroff2005novelty} was the first to explicitly explore the concept of Novelty Detection from texts. Under the paradigm of information retrieval, given a query, the TREC experiments were designed to retrieve relevant and novel sentences from a given collection. Some notable approaches for sentence level novelty detection from the TREC exercises are by 
\cite{allan2003retrieval,kwee2009sentence,li2005novelty,zhang2003expansion,collins2002information,gabrilovich2004newsjunkie,ru2004improved}. Textual Entailment based sentence level novelty mining was explored in the novelty subtask of RTE-TAC 6 and 7\cite{bentivogli2011seventh}. At the document level the problem is attempted by a few like \cite{zhang2002novelty,tsai2011d2s,karkali2013efficient,dasgupta2016automatic}. However we find that there is a dearth of a proper evaluation setup (e.g. corpus, baseline and evaluation methods) for novelty detection at the document level. This inspired us to create one and establish a benchmark for the same.
\subsection{Motivation and Contribution}
Our understanding and survey revealed that in spite of having several applications in various natural language processing (NLP) tasks, 
novelty detection at the document level has not attracted the coveted attention. Hence, we deem that novelty at the document level needs to be understood first, investigated in-depth, and benchmark setup (gold standard resources etc.) be created to validate the investigations as well as provide baselines for further research. 
We hope that the knowledge gained from this dataset and experiments would be a step towards our more ambitious vision of semantic level plagiarism detection in scholarly articles. Our contributions here could be outlined as :
\begin{itemize}
\item Proposing a benchmark dataset for document level novelty detection. We are unaware of the availability of any such corpus; and
\item A supervised machine learning model for document level novelty detection. This can be treated as a baseline model for further research. 
\end{itemize}
\section{Document Level Novelty}
Novelty detection from texts implies search for new information with respect to whatever is already known or seen.
Hence, the problem of novelty detection from texts is very subjective and depends upon the view of the intended reader. The knowledge of the reader regarding a particular event serves as the reference against which s/he decides the novelty of an incoming information. Careful observation of data characteristics led us to believe that 
\textit{Relevance}, \textit{Relativity} and \textit{Temporality} are three important properties of novelty detection. 
For example, searching for novelty between two documents, one talking about \textit{jaguar}, the animal and the other about \textit{jaguar}, the car is futile as one is not relevant to the other. Quite obvious that each one would contain different information than the other. 
Also when we talk about a document being novel it is always with respect to a reference set of documents already seen (information already gained from those seen documents) or what we say as the knowledge base of the reader. Also novel information is usually a temporal update over existing knowledge. With this view of novelty we went on to create a resource that effectively taps these properties, \textit{viz.,} \textbf{Relevance}, \textbf{Relativity} and \textbf{Temporality}. Our resource not only encompasses the lexical form of redundancy (a straight forward form of \textit{non-novelty}) but also delves deep into semantic textual redundancy (a more complex form of \textit{non-novelty}) with the expertise of human annotators.
\section{Benchmark Setup}
To address the issues pointed out in the previous section we develop a benchmark setup as discussed below. 
\subsection{Data Collection}
\begin{figure}[h!]
  \centering
  \includegraphics[width=0.5\textwidth,height= 4 cm]{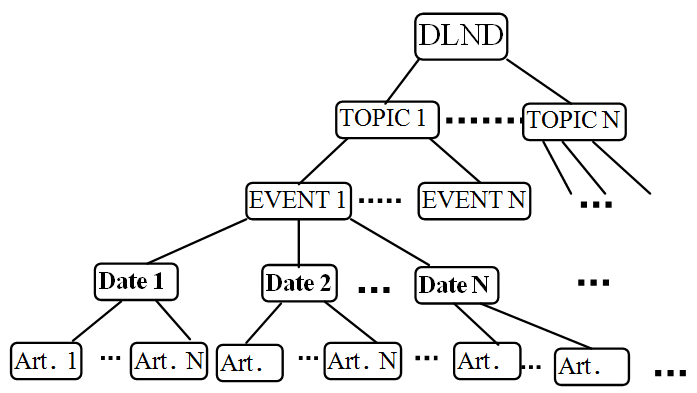}
  \caption{Temporal Crawling} 
  \label{dlnd_crawl}
\end{figure}

We design a web crawler\footnote{using the www.webhose.io API} to perform systematic, unbiased, \textit{event-specific} crawling of news articles, mostly from the online versions of Indian English newspapers. The news domains we looked into are : \textit{Accident (ACC), Politics (PLT), Business (BUS), Arts and Entertainment (ART), Crime (CRM), Nature (NAT), Terrorism (TER), Government (GOV), Sports (SPT), and Society (SOC)}. To ensure that \textit{Temporality} criteria is preserved, our web crawler is designed to fetch web documents for a certain event in a timely manner i.e. the crawled documents are grouped as per their dates of publications in different forums (See Figure \ref{dlnd_crawl}). Event wise statistics of the corpus are in Table \ref{DLND-cat}.
\begin{table}[h]
\begin{center}
\begin{tabular}{ |c|c| } 
\hline \bf Features & \bf Statistics \\ \hline
 Crawling period & Nov'16 - Nov'17  \\ \hline
 Number of events & 223  \\ \hline
 Number of sources per event & 3  \\ \hline 
 Total novel documents &  2736 \\ \hline
 Total non-novel documents & 2704   \\ \hline
 Total documents in TAP-DLND 1.0 & 6109  \\ \hline
 Average number of sentences &15\\\hline
 Average number of words&353\\\hline 
\end{tabular}
\end{center}
\caption{\label{DLND-stats} Statistics of TAP-DLND 1.0 corpus. Here, average number of sentences and words is per document.}
\end{table}
\subsection{Preprocessing}
As the data were crawled from various web sources\footnote{List of few news sources :
www.ndtv.com,
indianexpress.com,
timesofindia.indiatimes.com,
indiatoday.intoday.in,
thehindu.com,
news18.com,
firstpost.com,
dnaindia.com,
deccanchronicle.com,
financialexpress.com,
business-standard.com,
sify.com,
newskerala.com,
mid-day.com,
thedailystar.net,
theweek.in,
tribuneindia.com
} we perform some manual preprocessing works like removal of headlines, news source, date, time, noises (advertisements, images, hyperlinks) and convert the data into desired shape.

\subsection{Source Document Selection}
To mandate the \textit{Relevance} and \textit{Relativity} criteria, we select three documents for each event as the seed source documents. They are usually selected from the initial dates of reporting. Also so chosen that they represent different facets of information regarding that particular event (\textit{information coverage}). These source documents serve as the reference against which we asked the annotators to tag a target document (chosen from the remaining crawled documents for that event) as \textit{novel} or \textit{non-novel}. The source documents could be perceived as the memory of the reader or information already known against which it is to be determined with reasonable level of certainty that whether a target document contains sufficient new information to be labeled as \textit{novel}.
\begin{table}
\begin{center}
\begin{tabular}{ |c|c|c|c| } 
 \hline
 \bf Category & \bf \# Events & \bf \# N & \bf \# NN \\
 \hline
 \hline
 ACC & 10 & 231 & 272\\
 \hline
 PLT & 97 & 669 & 685\\
 \hline
 BUS & 35 & 202 & 264 \\ 
 \hline
 ART & 21 & 397 & 258\\ 
 \hline
 CRM & 10 & 237 & 174\\ 
 \hline
 NAT & 10 & 87 & 250 \\ 
 \hline
 TER & 18 & 255 & 468\\ 
 \hline
 GOV & 15 & 405 & 219\\ 
 \hline
 SPT & 2 & 39 & 51\\ 
 \hline
  SOC & 5 & 214 & 63\\ 
 \hline
\end{tabular}
\end{center}
\caption{\label{DLND-cat} Event wise statistics of TAP-DLND 1.0, $\# N\rightarrow$ Number of Novel documents, $\# NN\rightarrow$ Number of Non-Novel documents}
\end{table}
\subsection{Renaming files}For ease of information retrieval we rename each document in the corpus. A certain document bearing \textbf{'ACCE005SRC003.txt'} as file name indicates that it is the 3rd source document of the 5th event in the \textit{accident category}. For target documents 'SRC' is replaced by 'TGT'.
\subsection{Meta files} We generate meta files (.xml) for each document in the corpus. These meta files contain background information regarding a source/target document within structured XML tags and have the same file name as that of the corresponding document. The information content of the meta files are : \textit{date of publishing, publisher, title of reporting, source id, event id, event name, category, Document Level Annotation (DLA), number of words and sentences}. We develop a semi-automatic \textit{meta file generator interface} where attribute values are automatically captured from the hierarchically organized data (See Figure \ref{dlnd}). Stanford CoreNLP \cite{manning2014stanford} integrated with our interface gave us the field values for \textit{sentence} and \textit{word} count. We asked our annotators to provide their judgments for the \textit{DLA} attribute based on the guidelines specified in the next section.
\subsection{Annotation}
\begin{figure}[h!]
  \centering
  \includegraphics[width=0.5\textwidth,height=6 cm]{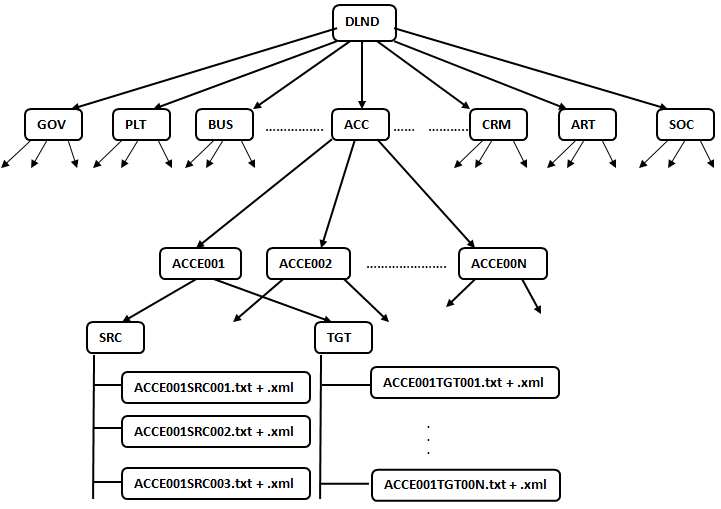}
  \caption{The DLND corpus structure} 
  \label{dlnd}
\end{figure}

Three annotators with post-graduate level knowledge in English were involved in labeling the DLND target documents. Having read the source document(s) we asked the annotators to annotate an incoming on-event document as \textit{non-novel} or \textit{novel} solely based on the information coverage in the source documents. The \textbf{annotation guidelines} were simple:
\begin{enumerate}
\item To annotate a document as \textit{non-novel} whose semantic content significantly overlaps with the source document(s) (maximum redundant information).
\item To annotate a document as \textit{novel} if its semantic content as well as intent (direction of reporting) significantly differs from the source document(s)(minimum or no information overlap). It could be an update on the same event or describing a post-event situation.
\item To leave out the ambiguous cases (for which the human annotators were not sure about the label).
\end{enumerate}
Two annotators independently labeled the target documents. The third annotator resolved the differences via majority voting. 
We found that novel items with respect to the source documents were mostly found in the reporting published in subsequent dates on the same event. Whereas non-novel items we found in the reporting published by different agency in the same date as that of the source documents. This is in line with the \textit{Temporality} criteria we discussed earlier. The inter-annotator agreement ratio was found to be \textbf{0.82} in terms of \textbf{Kappa coefficient}~\cite{fleiss1971measuring} which is assumed to be good as per ~\cite{landis1977measurement}.
The final structure of DLND is in Figure \ref{dlnd}.
\section{Evaluation}
\begin{figure}[h!]
  \centering
  \includegraphics[width=0.5\textwidth]{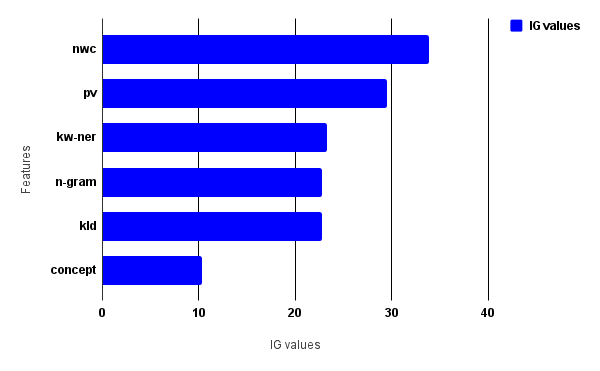}
  \caption{\label{feature significance} Significance of features \textit{based on Information Gain (IG)}. The length of the bar corresponds to the average merit (X : IG) of the feature (: Y).}
\end{figure}

\begin{savenotes}
\begin{table*}[ht]
\begin{center}
\begin{tabular}{ |c|c|p{11 cm}| }
\hline \bf Type & \bf Features & \bf Description\\
\hline  Semantic & Paragraph Vector (pv) + Cosine  & We represent the source and target documents in terms of \textit{paragraph vectors}\footnote{Distributed Bag-Of-Words (DBOW) paragraph vector model trained on Wikipedia articles.}\cite{le2014distributed}. Then we take the maximum of the cosine similarity between the source-target pairs.\\
\hline Semantic & Concept Centrality & To identify the central theme of a document we use the \textit{TextRank} summarization algorithm by \cite{mihalcea2004textrank}. Thereafter we vectorize the ranked summary for each source and target document by simple \textit{word2vec}\footnote{Trained on Google News Corpus of 100 billion words. 300 dimension vectors using CBOW model}\cite{mikolov2013distributed} concatenation. Finally we take the maximum of the cosine similarity between the source and target vectors. \\
\hline Lexical& n-gram similarity& We compute lexical overlap of target \textit{n-gram}'s with respect to source documents for $n$ = 2,3 and 8. Octagrams we use to put emphasis on phrase overlap.\\
\hline Lexical & Named Entities and & As Named Entities\footnote{Entities were extracted using the Stanford Tagger.} and Keywords\footnote{Using the Rapid Automatic Keyword Extraction (RAKE) algorithm.} play a significant role in determining\\ & Keywords match (kw-ner) & \textit{relevance}, we put additional weightage to them by considering their match (target w.r.t. sources) as a separate feature.\\
\hline Lexico- & New Word Count & The number of new words could be an effective indicator of the amount of\\  Semantic & (nwc) & novel information content in the target document w.r.t. the source(s) given. Here, for calculating new words, along with the surface forms, we consider their synonyms\footnote{Obtained from English WordNet~\cite{fellbaum1998wordnet}} as well to establish semantic relatedness.\\
\hline Language & Divergence& We use this feature to measure the dissimilarity between two documents\\ Model& (kld) &  represented as language models. We concatenate all the source documents into one and then measure the Kullback-Leibler Divergence with the target. \\
\hline
\end{tabular}
\end{center}
\caption{\label{feature_set} Feature Set}
\end{table*}
\end{savenotes}
We frame document level novelty detection as a binary classification problem and choose the features in parlance with the objective nature of texts that we consider for our experiments. We develop a binary classifier based on Random Forest\footnote{RF of 100 trees with minimum number of instances per leaf set to 1 implemented in WEKA machine learning toolkit} (RF)~\cite{Breiman2001} algorithm that classifies a document into either \textit{novel} or \textit{non-novel}. Our key focus is on extracting features that contribute to the semantics of a document. The set of features that we use for training and/or testing RF is listed in Table \ref{feature_set}. 
As is evident from the discussion in Section 3, TAP-DLND 1.0 consists a fair share of different levels (lexical as well as semantic) of text representations.
We first take a simple yet popular lexical baseline: Jaccard similarity with unigrams between the source document and the target \cite{zhang2003expansion}. We train a Logistic Regression (LR) classifier with the Jaccard score to classify a document based on its overlap with the source document. Table \ref{dlnd_res} clearly indicates that the lexical baseline fails miserably in identifying \textit{non-novel} documents. 
Next we went ahead with three approaches by \cite{zhang2002novelty} for novelty detection at the document level. The first one i.e. the Set Difference is essentially the count of new words in the target document with respect to the set of source document(s). For this we concatenate the source document(s) of each event to form one source against each target. The Geometric Distance measures the cosine similarity between two document vectors represented as \textit{tf-idf} vectors. For three source documents against one target document in TAP-DLND 1.0, we take the maximum of the cosine similarity score. The third approach measures the Kullback-Leibler divergence between the concatenated source document(s) and the prospective target document where a document $d$ is represented as a probabilistic unigram word distribution (language model $\theta_d$). Instead of setting a fixed threshold as \cite{zhang2003expansion}, we train a Logistic Regression classifier based on those measures to automatically determine the decision boundary. 
Another approach by \cite{karkali2013efficient} based on Novelty Scoring via Inverse Document Frequency (IDF) performed poorly in recognizing novel/non-novel documents in TAP-DLND 1.0. We also compare our method with a more recent approach of \cite{dasgupta2016automatic} on our data. This particular entropy-based approach produces novelty score ($NS$) of a
document $d$ with respect to a collection $C$. We
adapt their respective threshold criteria and infer
that documents with novelty score above (\textit{average+standard deviation}) are \textit{novel} and that with
novelty score below (\textit{average-standard deviation})
are \textit{non-novel}. We left out the remaining (average novelty class) cases for our experiments. Table \ref{dlnd_res} numbers clearly show that our method superseded the baselines and purported \textit{state-of-the-art} by a substantial margin. This significance we attribute to the choice of semantic features for our experiments (see Figure \ref{feature significance}). 
Lexico-Semantic feature new word count has the maximum contribution, 
for which we argue that novel events in context to newspaper articles would contain new entities, concepts, numbers whereas non-novel documents would consist identical or synonymous entities. Semantic features play a vital role which indicates that detection of novelty extends beyond lexical characteristics of text.
\begin{table*}[ht]
\begin{center}
\begin{tabular}{ |c|c|c|c|c|c|c|c| } 
\hline \bf Systems & \bf P(N) &  \bf R(N) & \bf $F_1$(N) & \bf P(NN) &  \bf R(NN) & \bf $F_1$(NN) & \bf Accuracy\\
 \hline Jaccard+LR (Baseline) & 52.2 & 96.1 & 67.6 & 74.0 & 10.9 & 19.0 & 53.8\\ 
\hline  Set Difference+LR & & & & & & &\\ \cite{zhang2002novelty} &  74.3 &  71.5 & 72.8 & 72.2 & 74.9 & 73.5 & 73.2\\ 
\hline  Geometric Distance+LR & & & & & & &\\ \cite{zhang2002novelty} &  65.6 &  84.3 & 73.7 & 84.2 & 55.3 & 66.7 & 69.8\\ 
\hline  Language Model (KLD)+LR & & & & & & &\\ \cite{zhang2002novelty} &  73.2 &  74.9 & 74.1 & 74.0 & 72.3 & 73.1 & 73.6\\ 
\hline  Novelty (IDF)+LR & & & & & & &\\ \cite{karkali2013efficient} &  52.5 &  92.1 & 66.9 & 66.5 & 15.9 & 25.6 & 54.2\\ 
 \hline \cite{dasgupta2016automatic}  & 65.1  & 63.8 & 64.4 & 64.1 & 65.3 & 64.6 & 64.5\\
 \hline \hline  \bf Proposed Approach (RF)  & \bf 77.6  &  82.3 & \bf 79.8 &  80.9 & \bf 76.1 & \bf 78.4 & 79.2\\ \hline
\end{tabular}
\end{center}
\caption{\label{dlnd_res} \textit{10-fold} cross-validation results on TAP-DLND 1.0 (in \%), $P\rightarrow Precision$, $R\rightarrow Recall$, $N\rightarrow Novel$, $NN\rightarrow Non-Novel$, $LR\rightarrow$ Logistic Regression,
$IDF\rightarrow$ Inverse Document Frequency,
$KLD\rightarrow$ Kullback- Leibler Divergence}
\end{table*}
\section{Conclusion}In this work we put forward a benchmark resource for \textit{document level novelty detection} and an evaluation scheme for the same. Our resource has an extensive coverage of ten different news categories and also includes the \textit{relevance, relativity,} and \textit{temporality} criteria inherently within its schema. Along with straightforward lexical characteristics it also manifests the high level semantic understanding of human annotators in its gold labels which is very essential for detecting semantic level redundancy. We hope that TAP-DLND 1.0 would evolve as a benchmark resource for experiments on document level novelty detection and provide valuable insights into the problem. In future we plan to annotate the TAP-DLND 1.0 corpus at the sentence level to have more fine perception regarding the amount of new information required to deem a document as \textit{novel}. Also we intend to include more target documents in data scarce categories.
\section{Acknowledgements}
The first author is supported by Visvesvaraya PhD Scheme, an initiative of Ministry of Electronics and Information Technology (MeitY), Government of India. The work is a product of the the Elsevier Centre of Excellence, Indian Institute of Technology Patna.


\section{Bibliographical References}
\label{main:ref}

\bibliographystyle{lrec}
\bibliography{xample}

\begin{thebibliography}{}

\bibitem[\protect\citename{Allan \bgroup et al.\egroup }1998]{allan1998line}
Allan, J., Papka, R., and Lavrenko, V.
\newblock (1998).
\newblock On-line new event detection and tracking.
\newblock In {\em Proceedings of the 21st annual international ACM SIGIR
  conference on Research and development in information retrieval}, pages
  37--45. ACM.

\bibitem[\protect\citename{Allan \bgroup et al.\egroup }2000]{allan2000first}
Allan, J., Lavrenko, V., and Jin, H.
\newblock (2000).
\newblock First story detection in tdt is hard.
\newblock In {\em Proceedings of the ninth international conference on
  Information and knowledge management}, pages 374--381. ACM.

\bibitem[\protect\citename{Allan \bgroup et al.\egroup
  }2003]{allan2003retrieval}
Allan, J., Wade, C., and Bolivar, A.
\newblock (2003).
\newblock Retrieval and novelty detection at the sentence level.
\newblock In {\em Proceedings of the 26th annual international ACM SIGIR
  conference on Research and development in informaion retrieval}, pages
  314--321. ACM.

\bibitem[\protect\citename{Allan}2002]{allan2002introduction}
Allan, J.
\newblock (2002).
\newblock Introduction to topic detection and tracking.
\newblock {\em Topic detection and tracking}, pages 1--16.

\bibitem[\protect\citename{Bentivogli \bgroup et al.\egroup
  }2011]{bentivogli2011seventh}
Bentivogli, L., Clark, P., Dagan, I., and Giampiccolo, D.
\newblock (2011).
\newblock The seventh pascal recognizing textual entailment challenge.
\newblock In {\em TAC}.

\bibitem[\protect\citename{Brants \bgroup et al.\egroup
  }2003]{brants2003system}
Brants, T., Chen, F., and Farahat, A.
\newblock (2003).
\newblock A system for new event detection.
\newblock In {\em Proceedings of the 26th annual international ACM SIGIR
  conference on Research and development in informaion retrieval}, pages
  330--337. ACM.

\bibitem[\protect\citename{Breiman}2001]{Breiman2001}
Breiman, L.
\newblock (2001).
\newblock Random forests.
\newblock {\em Machine Learning}, 45(1):5--32.

\bibitem[\protect\citename{Collins-Thompson \bgroup et al.\egroup
  }2002]{collins2002information}
Collins-Thompson, K., Ogilvie, P., Zhang, Y., and Callan, J.
\newblock (2002).
\newblock Information filtering, novelty detection, and named-page finding.
\newblock In {\em TREC}.

\bibitem[\protect\citename{Dasgupta and Dey}2016]{dasgupta2016automatic}
Dasgupta, T. and Dey, L.
\newblock (2016).
\newblock Automatic scoring for innovativeness of textual ideas.
\newblock In {\em Workshops at the Thirtieth AAAI Conference on Artificial
  Intelligence}.

\bibitem[\protect\citename{Fellbaum}1998]{fellbaum1998wordnet}
Fellbaum, C.
\newblock (1998).
\newblock {\em WordNet}.
\newblock Wiley Online Library.

\bibitem[\protect\citename{Fleiss}1971]{fleiss1971measuring}
Fleiss, J.~L.
\newblock (1971).
\newblock Measuring nominal scale agreement among many raters.
\newblock {\em Psychological bulletin}, 76(5):378.

\bibitem[\protect\citename{Franz \bgroup et al.\egroup }2001]{franz2001first}
Franz, M., Ittycheriah, A., McCarley, J.~S., and Ward, T.
\newblock (2001).
\newblock First story detection: Combining similarity and novelty based
  approaches.
\newblock In {\em Topic Detection and Tracking Workshop Report}, pages
  193--206.

\bibitem[\protect\citename{Gabrilovich \bgroup et al.\egroup
  }2004]{gabrilovich2004newsjunkie}
Gabrilovich, E., Dumais, S., and Horvitz, E.
\newblock (2004).
\newblock Newsjunkie: providing personalized newsfeeds via analysis of
  information novelty.
\newblock In {\em Proceedings of the 13th international conference on World
  Wide Web}, pages 482--490. ACM.

\bibitem[\protect\citename{Karkali \bgroup et al.\egroup
  }2013]{karkali2013efficient}
Karkali, M., Rousseau, F., Ntoulas, A., and Vazirgiannis, M.
\newblock (2013).
\newblock Efficient online novelty detection in news streams.
\newblock In {\em WISE (1)}, pages 57--71.

\bibitem[\protect\citename{Kwee \bgroup et al.\egroup }2009]{kwee2009sentence}
Kwee, A.~T., Tsai, F.~S., and Tang, W.
\newblock (2009).
\newblock Sentence-level novelty detection in english and malay.
\newblock In {\em Pacific-Asia Conference on Knowledge Discovery and Data
  Mining}, pages 40--51. Springer.

\bibitem[\protect\citename{Landis and Koch}1977]{landis1977measurement}
Landis, J.~R. and Koch, G.~G.
\newblock (1977).
\newblock The measurement of observer agreement for categorical data.
\newblock {\em biometrics}, pages 159--174.

\bibitem[\protect\citename{Le and Mikolov}2014]{le2014distributed}
Le, Q.~V. and Mikolov, T.
\newblock (2014).
\newblock Distributed representations of sentences and documents.
\newblock In {\em ICML}, volume~14, pages 1188--1196.

\bibitem[\protect\citename{Li and Croft}2005]{li2005novelty}
Li, X. and Croft, W.~B.
\newblock (2005).
\newblock Novelty detection based on sentence level patterns.
\newblock In {\em Proceedings of the 14th ACM international conference on
  Information and knowledge management}, pages 744--751. ACM.

\bibitem[\protect\citename{Manning \bgroup et al.\egroup
  }2014]{manning2014stanford}
Manning, C.~D., Surdeanu, M., Bauer, J., Finkel, J.~R., Bethard, S., and
  McClosky, D.
\newblock (2014).
\newblock The stanford corenlp natural language processing toolkit.
\newblock In {\em ACL (System Demonstrations)}, pages 55--60.

\bibitem[\protect\citename{Mihalcea and Tarau}2004]{mihalcea2004textrank}
Mihalcea, R. and Tarau, P.
\newblock (2004).
\newblock Textrank: Bringing order into texts.
\newblock Association for Computational Linguistics.

\bibitem[\protect\citename{Mikolov \bgroup et al.\egroup
  }2013]{mikolov2013distributed}
Mikolov, T., Sutskever, I., Chen, K., Corrado, G.~S., and Dean, J.
\newblock (2013).
\newblock Distributed representations of words and phrases and their
  compositionality.
\newblock In {\em Advances in neural information processing systems}, pages
  3111--3119.

\bibitem[\protect\citename{Ru \bgroup et al.\egroup }2004]{ru2004improved}
Ru, L., Zhao, L., Zhang, M., and Ma, S.
\newblock (2004).
\newblock Improved feature selection and redundance computing-thuir at trec
  2004 novelty track.
\newblock In {\em TREC}.

\bibitem[\protect\citename{Soboroff and Harman}2005]{soboroff2005novelty}
Soboroff, I. and Harman, D.
\newblock (2005).
\newblock Novelty detection: the trec experience.
\newblock In {\em Proceedings of the conference on Human Language Technology
  and Empirical Methods in Natural Language Processing}, pages 105--112.
  Association for Computational Linguistics.

\bibitem[\protect\citename{Stokes and Carthy}2001]{stokes2001first}
Stokes, N. and Carthy, J.
\newblock (2001).
\newblock First story detection using a composite document representation.
\newblock In {\em Proceedings of the first international conference on Human
  language technology research}, pages 1--8. Association for Computational
  Linguistics.

\bibitem[\protect\citename{Tsai and Zhang}2011]{tsai2011d2s}
Tsai, F.~S. and Zhang, Y.
\newblock (2011).
\newblock D2s: Document-to-sentence framework for novelty detection.
\newblock {\em Knowledge and information systems}, 29(2):419--433.

\bibitem[\protect\citename{Yang \bgroup et al.\egroup }1998]{yang1998study}
Yang, Y., Pierce, T., and Carbonell, J.
\newblock (1998).
\newblock A study of retrospective and on-line event detection.
\newblock In {\em Proceedings of the 21st annual international ACM SIGIR
  conference on Research and development in information retrieval}, pages
  28--36. ACM.

\bibitem[\protect\citename{Yang \bgroup et al.\egroup }2002]{yang2002topic}
Yang, Y., Zhang, J., Carbonell, J., and Jin, C.
\newblock (2002).
\newblock Topic-conditioned novelty detection.
\newblock In {\em Proceedings of the eighth ACM SIGKDD international conference
  on Knowledge discovery and data mining}, pages 688--693. ACM.

\bibitem[\protect\citename{Zhang \bgroup et al.\egroup }2002]{zhang2002novelty}
Zhang, Y., Callan, J., and Minka, T.
\newblock (2002).
\newblock Novelty and redundancy detection in adaptive filtering.
\newblock In {\em Proceedings of the 25th annual international ACM SIGIR
  conference on Research and development in information retrieval}, pages
  81--88. ACM.

\bibitem[\protect\citename{Zhang \bgroup et al.\egroup
  }2003]{zhang2003expansion}
Zhang, M., Song, R., Lin, C., Ma, S., Jiang, Z., Jin, Y., Liu, Y., Zhao, L.,
  and Ma, S.
\newblock (2003).
\newblock Expansion-based technologies in finding relevant and new information:
  Thu trec 2002: Novelty track experiments.
\newblock {\em NIST SPECIAL PUBLICATION SP}, (251):586--590.

\end{thebibliography}


\end{document}